\documentclass[runningheads]{llncs}

\usepackage[T1]{fontenc}
\usepackage{graphicx}
\usepackage{booktabs}
\usepackage{array}
\usepackage{amsmath}
\usepackage{xurl}
\usepackage{marvosym}
\usepackage{float}
\begin{document}

\title{TCM-DiagCommit: A TCM-SD-Derived Benchmark for Evidence-Conditioned Syndrome Selection and Commitment}
\titlerunning{TCM-DiagCommit}

\author{Erwei Hou\textsuperscript{*} \and
Yue Hu\textsuperscript{*\,(\Letter)}}

\authorrunning{E. Hou and Y. Hu}

\institute{Beijing University of Posts and Telecommunications, Beijing, China\\
\email{\{houerwei,huyue2023\}@bupt.edu.cn}}
\maketitle

\begingroup
\renewcommand{\thefootnote}{}
\footnotetext{\textsuperscript{*}Both authors contributed equally to this work.}
\endgroup

\begin{abstract}
Traditional Chinese Medicine (TCM) syndrome differentiation is usually evaluated as static final-label selection over complete clinical records. This setting hides three diagnostic behaviors: whether a model distinguishes semantic-neighbor syndrome labels from easy random distractors, whether it updates predictions as evidence fields are revealed, and whether it should commit under uncertainty. We introduce TCM-DiagCommit, a benchmark derived from the public TCM-SD/ZY-BERT resource while retaining its source syndrome labels; it converts 1,834 source cases into 23,842 task rows over 148 syndrome labels. The benchmark combines random, profile-neighbor, and full-vocabulary candidate ladders with staged t1/t2/t3 evidence. It also includes fixed-label tongue/pulse perturbations and insufficient-evidence variants. A benchmark-defined action contract scores commit, acquire, defer, and abstain. Across five zero-shot DirectLLM calibrations, random10 top-1 accuracy is 63.2\% to 66.4\%, but hard20 falls to 24.7\% to 30.3\% and full148 to 14.8\% to 22.3\%; on hard20, WT(all) wrong-terminal commits remain 62.2\% to 74.3\%. These results show that final-answer accuracy alone is insufficient for diagnostic benchmark evaluation: answer selection, evidence use, and benchmark commitment scoring should be measured separately.

\end{abstract}

\section{Introduction}

Medical QA and diagnosis benchmarks often cast diagnosis as complete-evidence final-answer selection. This formulation is easy to score, but it hides whether a model can revise an early hypothesis, move beyond semantic option elimination, or avoid committing when evidence is partial or conflicting. These hidden behaviors matter for evaluation even when the benchmark is not a clinical decision-support system.

TCM syndrome differentiation is a natural testbed for this problem. A syndrome pattern synthesizes chief complaint, history, inspection, and tongue/pulse evidence into a fine-grained diagnostic abstraction. TCM-SD made this task measurable with 54,152 real-world clinical records covering 148 syndromes and introduced the ZYBERT domain model \cite{ren2022tcmsd}, but the standard setting still presents a complete record and asks for one final label. We keep these source labels as gold labels and change the evaluation interface around them.

TCM-DiagCommit evaluates evidence-conditioned diagnostic decisions instead. We build a source-case-linked evaluation layer from TCM-SD/ZY-BERT \cite{ren2022tcmsd}: full-evidence candidate selection, t1/t2/t3 staged fields, random/profile-neighbor/full-vocabulary candidate ladders, fixed-label tongue/pulse perturbation controls, shortcut checks, and insufficient-evidence commitment tasks. The benchmark asks not only ``Which syndrome is most plausible?'' but also ``Did added evidence change the diagnosis?'' and ``Should the model commit now under the benchmark policy?''

We ask three research questions. \textbf{RQ1: Candidate difficulty.} Does hard20 expose semantic reranking pressure hidden by random10? \textbf{RQ2: Evidence.} Does staged evidence distinguish early guessing from evidence-corrected diagnosis? \textbf{RQ3: Commitment.} Do final-answer metrics miss commitment behavior under the same candidate set?

Our contributions follow four layers. \textbf{Formulation:} we recast TCM syndrome differentiation as evidence-conditioned diagnostic commitment. \textbf{Benchmark construction:} we provide 1,834 source cases, 148 syndrome labels, 26 task files covering 13 paired dev/test variants, and 23,842 rows with staged views, candidate ladders, fixed-label local perturbations, and insufficient-evidence variants. \textbf{Scoring protocol:} we define top-1/top-3 diagnosis, stage gains, wrong-terminal rates, expected cost, and source-case-level statistical reporting. \textbf{Empirical calibration:} five zero-shot DirectLLM runs reveal all three gaps, while a supervised ZY-BERT reference is used only to sanity-check benchmark learnability and difficulty calibration.

Resources are available at \url{https://github.com/Ethan-ai637/TCM-DiagCommit}.

\section{Related Work}

Medical NLP evaluation spans broad NLU suites, exam-style QA, clinical-case reasoning, and agentic diagnosis. Chinese resources such as CBLUE, CMB, and CMExam emphasize biomedical understanding or licensing-exam questions \cite{zhang2022cblue,wang2024cmb,liu2023cmexam}; they are valuable for knowledge and QA coverage, but they rarely expose the same source case under controlled evidence stages and candidate difficulty ladders. TCM-specific benchmarks such as TCMBench and MTCMB broaden TCM LLM evaluation beyond generic medical QA, covering knowledge and reasoning, with MTCMB further including prescription and safety-oriented tasks \cite{yue2024tcmbench,kong2025mtcmb}. TCM-DiagCommit instead provides a focused process-level evaluation within one 148-label syndrome ontology, measuring semantic-neighbor discrimination, evidence-conditioned updating, and terminal action reporting.

TCM-SD is the primary source, with 54,152 clinical records, 148 syndrome labels, and ZYBERT \cite{ren2022tcmsd}. Existing TCM syndrome-differentiation work includes supervised or knowledge-enhanced classification models such as dual-channel knowledge attention \cite{liu2025dcka}, as well as recent work evaluating and improving LLMs' syndrome-differentiation thinking ability \cite{chen2025syndrome}. These methods generally optimize final-label prediction or generated reasoning. Our benchmark keeps the final-label task but adds protocol-level stressors around it: profile-neighbor candidates test semantic reranking, staged rows test field-conditioned updating, and commitment rows test action reporting under a fixed scorer.

Dynamic medical evaluation provides the second thread. Sequential diagnosis and agentic clinical benchmarks motivate additional-evidence and tool-use policies: Sequential Diagnosis evaluates stepwise diagnostic encounters \cite{nori2025sequential}, while AgentClinic evaluates simulated patient interaction, multimodal data collection, and tool use \cite{schmidgall2026agentclinic}. TCM-DiagCommit deliberately does not simulate open dialogue: it uses fixed field masking so every stage and action is paired across source cases. Selective classification formalizes risk-coverage trade-offs, while conformal prediction provides coverage-guaranteed uncertainty sets \cite{geifman2017selective,angelopoulos2021conformal}; calibration work shows confidence can misalign with correctness \cite{guo2017calibration}. Counterfactually augmented NLP data test shortcut reliance \cite{kaushik2020counterfactual}; our fixed-gold tongue/pulse rows follow that stress-test spirit while avoiding claims that the perturbed row has a new clinical gold label.
Table~\ref{tab:related-compare} summarizes the positioning across evaluation capabilities.

\begin{table}
\caption{Benchmark comparison by evaluation capabilities.}
\label{tab:related-compare}
\centering
\begin{tabular*}{\textwidth}{@{\extracolsep{\fill}}lccccc@{}}
\toprule
Benchmark & Staged & Hard & Insuff. & Actions & Perturb. \\
\midrule
TCM-SD/ZY-BERT & No & No & No & No & No \\
Chinese medical QA & No & No & Partial & No & No \\
TCM LLM benchmarks & Partial & No & Partial & No & Partial \\
Sequential/agentic diagnosis & Yes & Task-specific & Partial & Yes & Partial \\
\textbf{TCM-DiagCommit} & \textbf{Yes} & \textbf{Yes} & \textbf{Yes} & \textbf{Yes} & \textbf{Yes} \\
\bottomrule
\end{tabular*}
\end{table}

\section{Benchmark Construction}

\subsection{Source and Release Boundary}

The core source is TCM-SD/ZY-BERT \cite{ren2022tcmsd}. We preserve the source syndrome labels from TCM-SD and change the evaluation interface around them. The benchmark release contains derived task rows, source-case links, and release metadata for this TCM-SD-derived evaluation layer. Auxiliary TCM QA, dialogue, or mixed-task resources are used only as related-work context. TCM-SD train records build only hard-candidate label profiles; held-out evaluation source cases are selected from the source dev/test splits.

Formally, a selected source case is $x_i=(f_i^{(1)},f_i^{(2)},f_i^{(3)},y_i)$, where $f_i^{(1)}$ is chief complaint, $f_i^{(2)}$ is description/history, $f_i^{(3)}$ is detection/tongue-pulse evidence when present, and $y_i\in\mathcal{Y}$ with $|\mathcal{Y}|=148$. Stage views are $E_i^{(1)}=f_i^{(1)}$, $E_i^{(2)}=[f_i^{(1)};f_i^{(2)}]$, and $E_i^{(3)}=[f_i^{(1)};f_i^{(2)};f_i^{(3)}]$. For candidate policy $\pi$, $C_i^\pi$ always contains $y_i$. A model output at stage $s$ is $o_i^{(s)}=(\hat y_i^{(s)},T_i^{(s)},c_i^{(s)},a_i^{(s)})$: top-1, top-3, confidence, and action.

\begin{figure}[H]
\centering
\includegraphics[width=\textwidth]{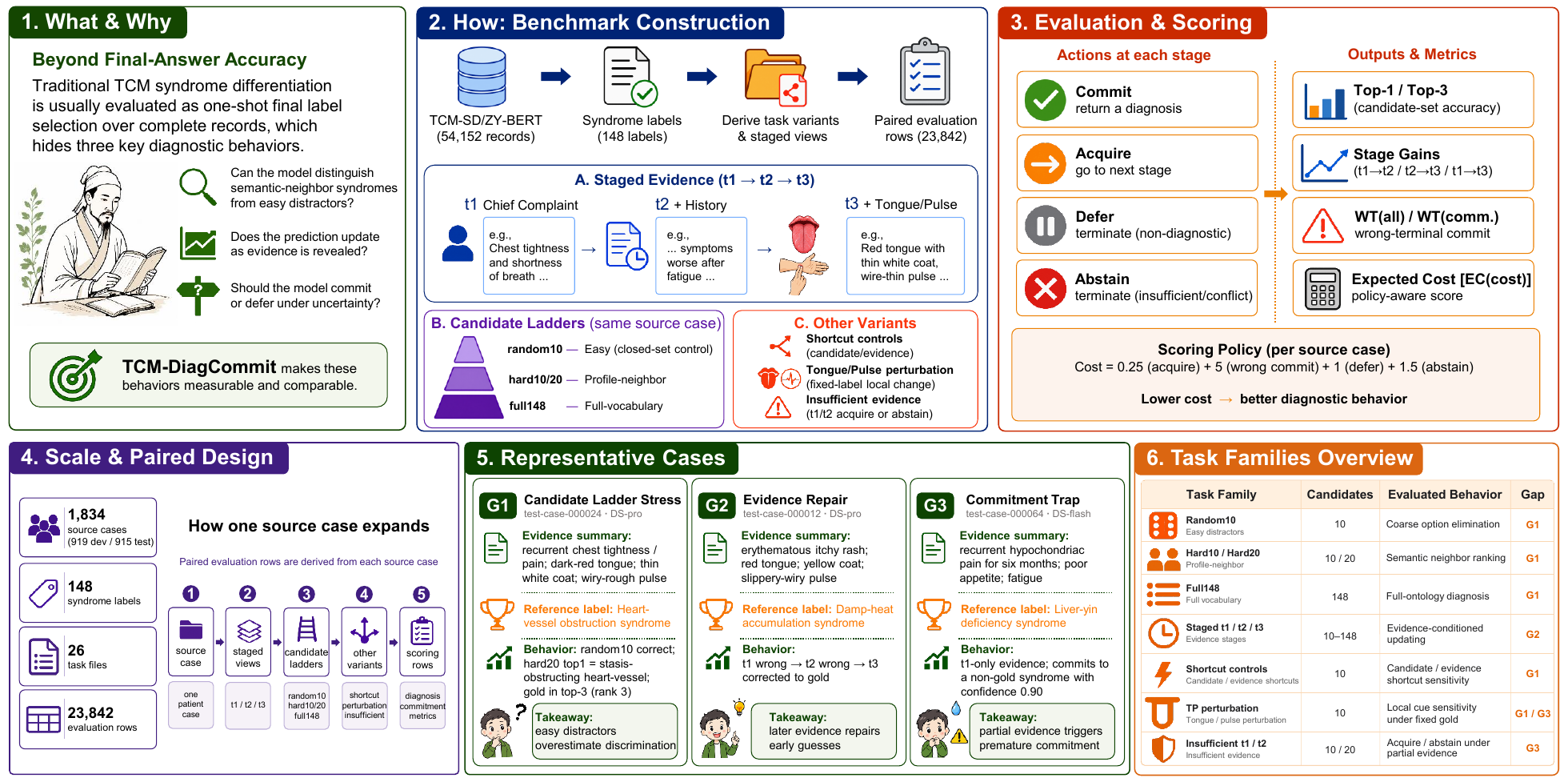}
\caption{Benchmark overview. A source case is shown through staged evidence and paired task variants; TCM-DiagCommit evaluates semantic-neighbor discrimination, evidence updating, and diagnostic commitment behavior with source-case-level scoring.}
\label{fig:overview}
\end{figure}

The release contains 919 dev cases and 915 test cases; both splits cover all 148 labels, with 2--10 selected cases per label per split. The cap keeps the paired variant matrix tractable while preserving label coverage and near-identical dev/test balance. Table~\ref{tab:dataset-scale} and Figure~\ref{fig:overview} summarize the scale: 26 task files and 23,842 task rows.

\noindent\textbf{Construction recipe.}
The deterministic pipeline (1) normalizes dev/test rows, stratifies by syndrome, caps each label for balanced held-out coverage, removes exact and near-duplicate cross-split cases, and preserves source-case ids; (2) builds train-only label profiles from Chinese character n-gram vectors over disease and clinical fields; (3) ranks label neighbors by cosine similarity, $\mathrm{sim}(a,b)=p_a^\top p_b/(\|p_a\|_2\|p_b\|_2)$, then materializes \texttt{random10}, \texttt{hard10}, \texttt{hard20}, and \texttt{full148} candidate sets with shuffled gold slots; and (4) emits static, staged, shortcut, fixed-label tongue/pulse perturbation, and insufficient-evidence rows. Perturbation rows choose a same-split non-gold donor with extractable tongue/pulse evidence, replace the source tongue/pulse phrase, keep the source gold fixed, and inject the donor label into hard10 candidates. Thus ``hard'' denotes reproducible profile-neighbor alternatives selected by the same train-only similarity recipe: it creates semantic-neighbor ranking pressure rather than an extra source of labels.

\noindent\textit{Complexity.} Profile extraction is $O(N_{\mathrm{train}}L)$, label-neighbor precomputation is $O(|\mathcal{Y}|^2d)$, per-row candidate lookup is $O(k)$, commitment scoring is $O(NS)$ for $S\le3$, and bootstrap reporting is $O(BN)$ for $B$ source-case resamples.

\begin{table}
\caption{TCM-DiagCommit release scale. Each source case is transformed into paired task variants; dev and test share the same 148-label ontology but have disjoint cases. The statistical unit is source case, not task row.}
\label{tab:dataset-scale}
\centering
\begin{tabular*}{\textwidth}{@{\extracolsep{\fill}}lrrrrrr@{}}
\toprule
Split & Source cases & Labels & Min/label & Max/label & Task files & Task rows \\
\midrule
Dev & 919 & 148 & 2 & 10 & 13 & 11,947 \\
Test & 915 & 148 & 2 & 10 & 13 & 11,895 \\
\midrule
Total & 1,834 & 148 & 4 & 20 & 26 & 23,842 \\
\bottomrule
\end{tabular*}
\end{table}

\subsection{Task Variants}

Table~\ref{tab:variant-families} summarizes the benchmark families. Each task file contains one row per selected source case (919 dev, 915 test). The candidate ladder is central: \texttt{random10} is an easy closed-set control; \texttt{hard10}/\texttt{hard20} use profile-neighbor distractors; and \texttt{full148} evaluates the full vocabulary. Gold is always present, so failure is a diagnosis or ranking error rather than an open-vocabulary mismatch. We use G1 for semantic candidate discrimination, G2 for staged evidence integration, and G3 for commitment behavior.

\begin{table}
\caption{Task families in the public release. G1 = semantic candidate discrimination, G2 = staged evidence integration, G3 = benchmark commitment behavior.}
\label{tab:variant-families}
\centering
\begin{tabular*}{\textwidth}{@{\extracolsep{\fill}}>{\raggedright\arraybackslash}p{0.28\textwidth}c>{\raggedright\arraybackslash}p{0.34\textwidth}c@{}}
\toprule
Task family & Candidates & Evaluated behavior & Gap \\
\midrule
Random10 & 10 & Coarse option elimination. & G1 \\
Hard10/20 & 10/20 & Semantic profile-neighbor reranking. & G1 \\
Full148 & 148 & Full-ontology diagnosis. & G1 \\
Staged t1/t2/t3 & 10--148 & Evidence-conditioned updating. & G2 \\
Shortcut controls & 10 & Candidate/evidence shortcut sensitivity. & G1 \\
TP perturbation & 10 & Local cue sensitivity under fixed gold. & G1/G3 \\
Insufficient t1/t2 & 10/20 & Acquire/abstain under partial evidence. & G3 \\
\bottomrule
\end{tabular*}
\end{table}

\subsection{Progressive Evidence}

TCM-DiagCommit uses a field-level staged evidence protocol designed with reference to the information flow of TCM consultation: an initial complaint motivates questioning, history contextualizes the condition, and inspection plus tongue/pulse evidence refine syndrome differentiation. The t1/t2/t3 reveal policy is therefore a diagnostic-workflow-informed staged-information abstraction for benchmark evaluation. Stage t1 exposes chief complaint, t2 adds description/history, and t3 adds detection, including inspection and tongue/pulse-style observations when present. The split is deterministic evidence masking with no generated dialogue or summary, so each row remains traceable to one source case. Predictions are scored at every stage to measure whether added fields correct early errors or cause label flips.

\subsection{Diagnostic Commitment}

For commitment evaluation, each prediction includes one scored action at the current stage: \texttt{commit} returns a diagnosis; \texttt{acquire} advances the scorer to the next evidence stage; \texttt{defer} returns a terminal non-diagnostic outcome; and \texttt{abstain} returns a terminal no-diagnosis outcome under insufficient or conflicting evidence. The scorer moves from t1 to t3 and stops at the first terminal action. Under the task-family scoring policy, full-evidence and no-tongue/pulse variants use a commit-oriented target, whereas rows exposing only candidates, a chief complaint, tongue/pulse evidence, or partial t1/t2 evidence use an acquire-or-abstain target. Tongue/pulse perturbation rows are evaluated as fixed-label cue-sensitivity controls. These family-level targets provide a consistent scoring interface across rows.

Expected~cost uses fixed penalties per source case. Defaults are 0.25 for an evidence step, 5 for a wrong commit, 1 for defer, and 1.5 for abstain:
\[
\begin{aligned}
\mathrm{Cost}_i&=\lambda_s(s_i^\star-1)+\lambda_w W_i+\lambda_d \mathbf{1}[a_i^\star=\mathrm{defer}]
+\lambda_a \mathbf{1}[a_i^\star=\mathrm{abstain}],\\
W_i&=\mathbf{1}[a_i^\star=\mathrm{commit}\land \hat y_i^\star\ne y_i].
\end{aligned}
\]
Wrong-terminal rates are reported both over all cases and over committed cases:
\[
\mathrm{WT(all)}=N^{-1}\sum_i W_i,\qquad
\mathrm{WT(comm.)}=\frac{\sum_i W_i}{\sum_i \mathbf{1}[a_i^\star=\mathrm{commit}]}.
\]
For manuscript reporting, WT(all) is the all-case wrong-terminal rate, while WT(comm.) is the corresponding rate among committed cases. The pre-t3 commit rate records any commit before t3. WT rates are determined directly from model outputs, while expected cost summarizes the same outputs under the stated penalty weights.

\section{Evaluation Protocol}

\subsection{Prediction Contract}

All methods must output JSON-compatible predictions with a stable method id, variant, case id, optional stage, \texttt{candidate\_syndromes}, \texttt{top1}, \texttt{top3}, confidence, action, and \texttt{need\_more\_evidence}. For closed-set variants, top-1 and top-3 labels must come from the provided candidates. Invalid, unparsable, or out-of-candidate predictions are counted explicitly rather than repaired silently.

For reproducibility and provider comparability, DirectLLM uses one fixed two-message prompt template, \texttt{tcm\_commitbench\_v1}, and JSON parser for every model; protocol files show row-to-prompt and prediction-to-scorer conversion. The system message states that the task is a research benchmark, not clinical advice; labels must be copied from candidates; output must be strict JSON; wrong high-confidence answers are benchmark errors; and missing symptoms should not be invented. The user JSON contains \texttt{task}, \texttt{evidence\_stage}, \texttt{case\_id}, visible evidence, candidates, output schema, and copying rules; no provider receives extra provider-specific instructions or repair prompts. Commitment actions are \texttt{commit}, \texttt{acquire}, \texttt{defer}, or \texttt{abstain}. Holding this prompt surface fixed makes model and task-variant comparisons directly paired.

\subsection{Metrics}

We organize metrics around the three evaluation gaps rather than treating every logged field as a main result. Macro-F1, invalid rate, token use, accuracy per 1k tokens, overconfidence error, and full action distributions are retained as diagnostic artifact metrics but are not used as headline claims.

G1 uses random/hard/full top-1 and top-3 to assess semantic candidate discrimination; G2 uses t1/t2/t3 top-1, t3--t1 gain, and prediction flips to assess staged evidence integration; G3 uses commit rate, WT(all), WT(comm.), premature commit, and expected cost to assess commitment behavior. Additional metrics are retained for secondary analysis, keeping the primary claims focused.

Non-model reference metrics are computed separately from model performance. On the held-out test split, uniform top-1 chance is 10.0\% for random10 and hard10, 5.0\% for hard20, and 0.7\% for full148. Uniform top-3 chance is 30.0\%, 15.0\%, and 2.0\%, respectively. Candidate-first top-1 accuracy is close to chance for hard20 (5.0\%) and below chance for full148 (0.3\%), indicating no strong first-position shortcut. These references are used only to calibrate difficulty and order bias; they are not reported as system results.

Statistical reporting is fixed before model comparison: source-case bootstrap confidence intervals within each split and variant, paired deltas for shared source-case cells, and McNemar or paired sign tests as supporting evidence. Main result tables report 95\% intervals for the headline DirectLLM metrics, while artifact CSVs retain the full metric set. Intervals use 10,000 source-case bootstrap resamples; paired deltas use the same source-case ids across variants. This design preserves case-level comparability across analyses. We report five complete DirectLLM test matrices under this locked protocol.

\subsection{DirectLLM Calibration Setup}

The reported calibration uses DirectLLM: each model receives the benchmark prompt and returns the required JSON. Static, progressive, and commitment scores use task-specific contracts, so static full-evidence accuracy and progressive t3 accuracy are parallel calibration points. Commitment rows use the same source case, candidate set, and visible stage as matched diagnosis rows, but action and top-1 are generated together under a commitment-specific prompt; Table~\ref{tab:behavior-models} is paired task evidence, not a natural conversation trajectory. Runs use temperature 0 and omit the provider \texttt{max\_tokens} request field; results are timestamped API snapshots. No other zero-shot or policy method is treated as a competing system; the scorer itself is model-agnostic. This locked setup keeps provider comparisons aligned under the same prompting and scoring conditions.

The completed set includes DeepSeek-v4-flash, DeepSeek-v4-pro, MiniMax-M3, Mimo-v2.5-pro, and Qwen-3.7-Plus, each with static, progressive, and commitment outputs over the 915 held-out test cases across the locked random10, hard10, hard20, full148, perturbation, and insufficient-evidence variants. Incomplete provider runs are excluded.

\begin{figure}
\centering
\includegraphics[width=0.94\textwidth]{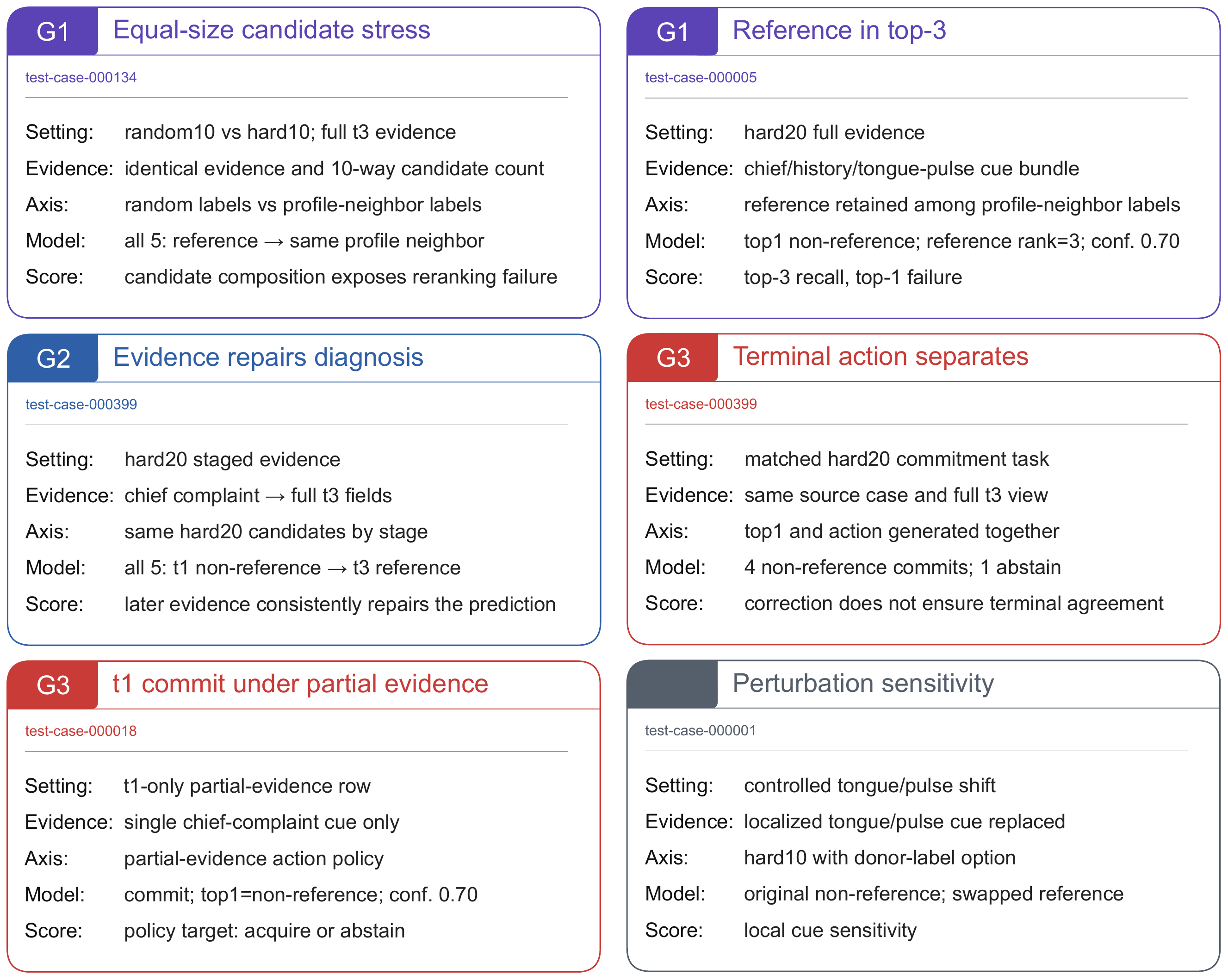}
\caption{Representative source-linked cases across the three evaluation gaps.}
\label{fig:case-grid}
\end{figure}
\section{Results}

\subsection{Sanity Reference: Learnability and Difficulty Calibration}

We include a task-specific ZY-BERT reference only as a benchmark sanity check, not as a DirectLLM comparator. It is fine-tuned with TCM-SD train labels and uses the source development labels for model selection; the reported numbers are evaluated only on the held-out benchmark test split. Candidate-masked probabilities show learnable signal and difficulty calibration: random10 87.8, hard20 61.3, full148 58.8; on hard20, t1/t2/t3 are 26.3/56.1/61.3. The same directional random-to-hard decrease and t1-to-t3 increase therefore appear in both this supervised reference and the zero-shot DirectLLM results. ZY-BERT does not evaluate a commit/acquire/defer/abstain policy.

The larger random10-to-hard20 drop locates most supervised ladder difficulty in profile-neighbor composition. The stage pattern also confirms that history and detection fields add signal beyond the chief complaint. This supervised reference separates benchmark learnability from behavior induced by the zero-shot prompting interface.

\subsection{Gap 1: Random Distractors Overestimate Semantic Discrimination}

\begin{figure}[H]
\centering
\includegraphics[width=0.78\textwidth]{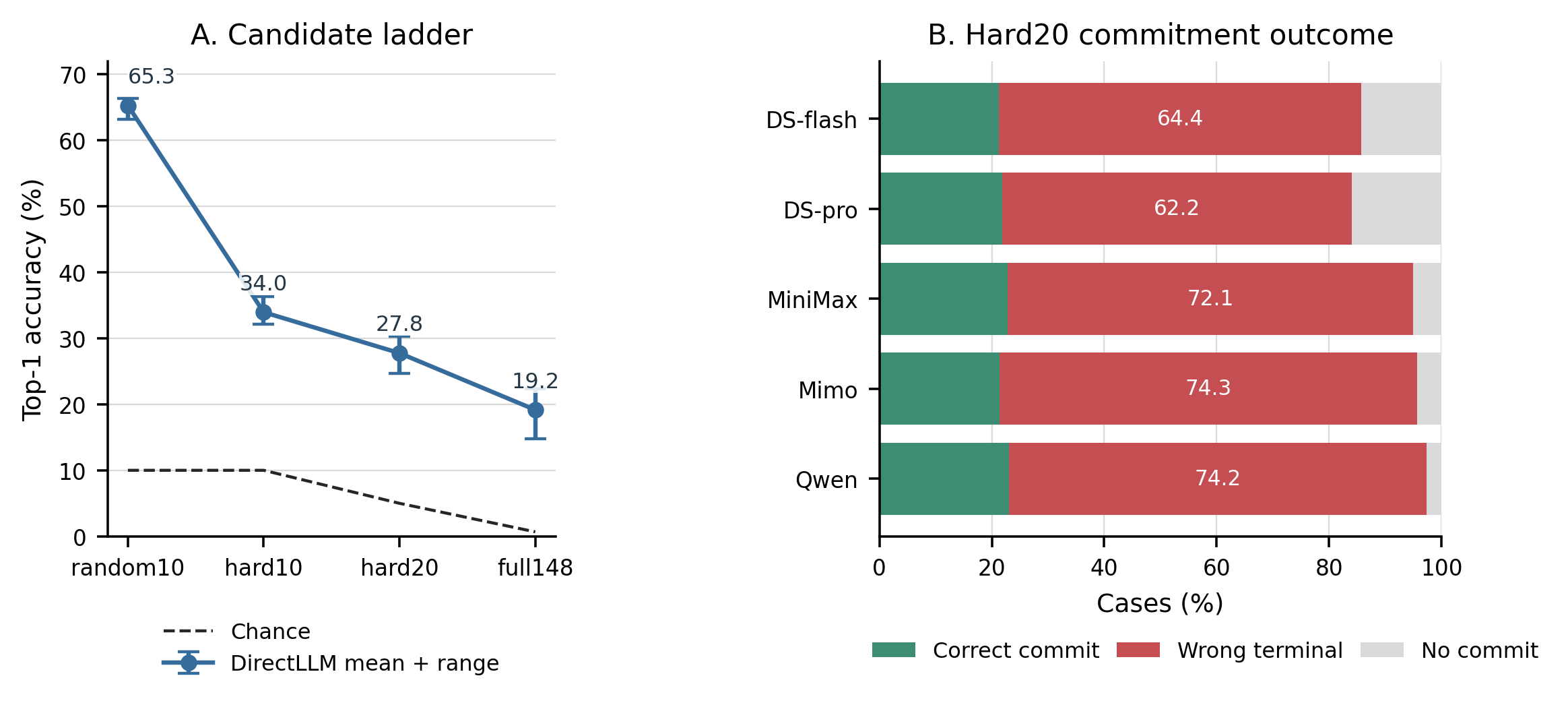}
\caption{Five DirectLLM calibration runs. Panel A shows mean/min-max static top-1 with chance baselines; Panel B shows hard20 terminal outcomes with WT(all) labels.}
\label{fig:directllm-summary}
\end{figure}

Table~\ref{tab:static-models} reports static top-1 accuracy with 95\% source-case bootstrap intervals across all four candidate regimes. Random10 reaches 63.2\%--66.4\%, hard10 falls to 32.1\%--36.3\%, hard20 to 24.7\%--30.3\%, and full148 to 14.8\%--22.3\%. Thus random distractors reward coarse semantic elimination, whereas the hard settings stress ranking among train-profile-neighbor labels. Figure~\ref{fig:directllm-summary}A shows the same ladder.

All five models follow the same monotonic pattern. Because random10 and hard10 both contain ten candidates, their 30.1--33.3 point gap isolates candidate composition from candidate count. Expanding hard10 to hard20 produces a further 5.3--7.4 point decrease, showing that both semantic proximity and denser neighbor competition contribute to difficulty.

\begin{table}
\caption{Zero-shot DirectLLM diagnosis across the complete candidate ladder. Values are top-1 accuracy (\%) with 95\% source-case bootstrap CI.}
\label{tab:static-models}
\centering
\setlength{\tabcolsep}{2.1pt}
\begin{tabular*}{\textwidth}{@{\extracolsep{\fill}}lcccc@{}}
\toprule
Model & random10 & hard10 & hard20 & full148 \\
\midrule
DS-flash & 65.5 [62.4,68.5] & 34.4 [31.3,37.5] & 28.6 [25.7,31.5] & 17.7 [15.3,20.2] \\
DS-pro & 65.9 [62.8,69.0] & 34.9 [31.8,37.9] & 29.6 [26.7,32.7] & 21.5 [18.9,24.2] \\
MiniMax & 66.4 [63.4,69.5] & 36.3 [33.1,39.3] & 30.3 [27.3,33.2] & 22.3 [19.7,25.0] \\
Mimo & 63.2 [60.0,66.2] & 32.1 [29.2,35.2] & 25.6 [22.8,28.4] & 14.8 [12.5,17.0] \\
Qwen & 65.4 [62.2,68.5] & 32.1 [29.2,35.3] & 24.7 [22.0,27.5] & 19.5 [16.8,22.0]
\\
\bottomrule
\end{tabular*}
\end{table}

\subsection{Gap 2: Staged Evidence Supports Partial Updating}

\begin{figure}
\centering
\includegraphics[width=0.82\textwidth]{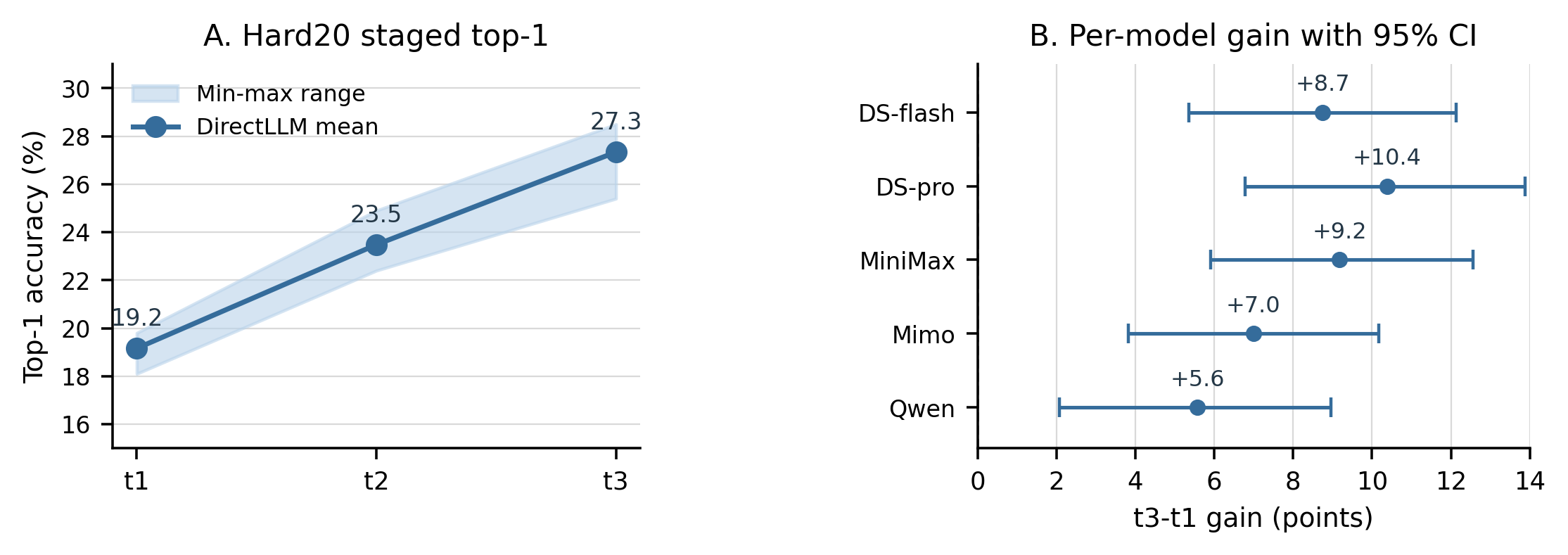}
\caption{Hard20 staged evidence curve and gain analysis. Panel A shows DirectLLM mean top-1 with min-max range across t1/t2/t3; Panel B shows each model's t3--t1 gain with 95\% source-case bootstrap CI.}
\label{fig:staged-curve}
\end{figure}

Figure~\ref{fig:staged-curve} shows hard20 progressive top-1 accuracy from t1 to t3 and per-model t3--t1 gains. Mean accuracy rises from 19.2\% at t1 to 27.3\% at t3; Table~\ref{tab:behavior-models} reports the same gains with 95\% CI. The gains range from +5.6 to +10.4 points, and every interval excludes zero. The curve supports the staged-evidence premise: later fields add useful signal. At the same time, t3 remains far below solved, so the result is partial evidence integration rather than strong final diagnosis.

\subsection{Gap 3: Commitment Scoring Exposes Wrong-Terminal Behavior}

Table~\ref{tab:behavior-models} groups hard20 staged evidence and commitment metrics with 95\% CIs. The gain column is from progressive rows; commitment columns are from matched hard20 commitment rows where \texttt{top1}, confidence, and action are generated together. Commit rates are 84.0\%--97.3\%, and WT(all) is 62.2\%--74.3\%. WT(comm.) is 74.0\%--77.6\%, while cost is 3.61--4.06 per source case. These results show that frequent commitment can coexist with high wrong-terminal rates. Figure~\ref{fig:directllm-summary}B shows why final-answer accuracy and diagnostic commitment must be scored separately.

\begin{table}
\caption{Hard20 DirectLLM behavior with 95\% CI. Gain is progressive t3--t1; commitment columns come from matched hard20 commitment rows in which top-1 and action are generated together. WT(all) = wrong-terminal commit over all cases; WT(comm.) = wrong-terminal commit among committed cases.}
\label{tab:behavior-models}
\centering
\setlength{\tabcolsep}{1.6pt}
\begin{tabular*}{\textwidth}{@{\extracolsep{\fill}}lccccc@{}}
\toprule
Model & Gain & Commit & WT(all) & WT(comm.) & Cost \\
\midrule
DS-flash & \scriptsize 8.7 [5.4,12.1] & \scriptsize 85.7 [83.4,87.9] & \scriptsize 64.4 [61.2,67.4] & \scriptsize 75.1 [72.0,78.2] & \scriptsize 3.65 [3.51,3.80] \\
DS-pro & \scriptsize 10.4 [6.8,13.9] & \scriptsize 84.0 [81.6,86.3] & \scriptsize 62.2 [59.0,65.2] & \scriptsize 74.0 [70.9,77.0] & \scriptsize 3.61 [3.46,3.74] \\
MiniMax & \scriptsize 9.2 [5.9,12.6] & \scriptsize 94.9 [93.4,96.2] & \scriptsize 72.1 [69.2,75.0] & \scriptsize 76.0 [73.2,78.8] & \scriptsize 3.90 [3.76,4.03] \\
Mimo & \scriptsize 7.0 [3.8,10.2] & \scriptsize 95.7 [94.4,97.0] & \scriptsize 74.3 [71.4,77.2] & \scriptsize 77.6 [74.8,80.3] & \scriptsize 4.06 [3.92,4.19] \\
Qwen & \scriptsize 5.6 [2.1,9.0] & \scriptsize 97.3 [96.2,98.3] & \scriptsize 74.2 [71.5,77.0] & \scriptsize 76.3 [73.6,79.1] & \scriptsize 4.00 [3.87,4.14]
\\
\bottomrule
\end{tabular*}
\end{table}

Table~\ref{tab:insufficient-models} gives the full insufficient-evidence t1/t2 action summary. These rows further separate action policy from diagnosis accuracy: t1 commit rates range from 1.2\% to 13.7\%, while t2 commit rates rise to 45.4\%--81.2\%. The benchmark can therefore express different partial-evidence policies without changing the closed-set label candidates.

\begin{table}
\caption{Insufficient-evidence hard20 commitment behavior. Values are percentages with 95\% source-case bootstrap CI.}
\label{tab:insufficient-models}
\centering
\setlength{\tabcolsep}{1.8pt}
\begin{tabular*}{\textwidth}{@{\extracolsep{\fill}}lcccc@{}}
\toprule
Model & t1 Commit & t1 WT(all) & t2 Commit & t2 WT(all) \\
\midrule
DS-flash & 9.2 [7.3,11.1] & 6.0 [4.5,7.5] & 67.2 [64.0,70.2] & 48.7 [45.5,52.0] \\
DS-pro & 1.2 [0.5,2.0] & 0.8 [0.2,1.4] & 45.4 [42.2,48.6] & 33.8 [30.8,36.7] \\
MiniMax & 13.7 [11.4,16.0] & 10.6 [8.6,12.6] & 77.7 [75.0,80.3] & 58.9 [55.7,62.1] \\
Mimo & 6.2 [4.7,7.9] & 4.5 [3.2,5.9] & 81.2 [78.7,83.7] & 61.2 [58.0,64.4] \\
Qwen & 4.2 [3.0,5.5] & 3.4 [2.3,4.6] & 70.2 [67.1,73.1] & 53.7 [50.4,56.9]
\\
\bottomrule
\end{tabular*}
\end{table}

\subsection{Failure Decomposition and Qualitative Cases}

Together, Figure~\ref{fig:case-grid} and Table~\ref{tab:failure-decomposition} summarize paired row-level results as a compact failure taxonomy. The identified patterns connect the three gaps: staged evidence can repair early guesses, candidate recall can occur without successful top-1 reranking, and a correct staged prediction can still be followed by a wrong terminal commitment under the separate commitment prompt. These patterns provide a concise row-level view of the three evaluation gaps. The figure and table complement the aggregate results with representative failure cases.

\begin{table}
\caption{Failure decomposition over five hard20 DirectLLM runs. Counts are ranges over models out of 915 held-out source cases.}
\label{tab:failure-decomposition}
\centering
\setlength{\tabcolsep}{4pt}
\begin{tabular*}{\textwidth}{@{\extracolsep{\fill}}>{\raggedright\arraybackslash}p{0.31\textwidth}>{\raggedright\arraybackslash}p{0.17\textwidth}>{\raggedright\arraybackslash}p{0.42\textwidth}@{}}
\toprule
Failure bucket & Range & Interpretation \\
\midrule
t1 wrong -> t3 correct & 137--187 & Added fields can repair early guesses. \\
Correctable but wrong commit & 58--81 & The action trajectory can fail despite a correct t3 prediction. \\
Static wrong, gold in top-3 & 209--264 & Candidate recall can exist while reranking fails.
\\
\bottomrule
\end{tabular*}
\end{table}

\section{Discussion}

The results reveal three distinct bottlenecks: profile-neighbor reranking, incomplete integration of additional evidence, and disagreement between terminal actions and inherited labels. Holding source cases fixed isolates these effects while candidate composition, evidence stage, and action scoring vary. The equal-size random10/hard10 example in Figure~\ref{fig:case-grid} separates candidate composition from candidate count, and the shared case 000399 shows that a t3 correction need not carry over to the matched terminal action. G1 therefore measures sensitivity to train-profile semantic proximity, while G3 measures terminal agreement under the locked benchmark policy. Fixed-label tongue/pulse rows complement these axes by measuring prediction response to a localized cue change.

\noindent\textbf{Limitations and future validation.}
TCM-DiagCommit inherits the data scope of TCM-SD/ZY-BERT and uses label-balanced sampling, so its label distribution may differ from real-world syndrome prevalence. Source-case intervals quantify case variation under the locked prompt; prompt wording and evolving API snapshots remain separate robustness dimensions. Future work can add clinician stage-wise anchors, alternative neighbor constructions, supervised retrieval/reranking baselines, and calibrated commitment policies.

\section{Conclusion}

TCM-DiagCommit reframes TCM syndrome differentiation from static label selection into evidence-conditioned syndrome selection and benchmark commitment scoring while preserving TCM-SD source labels. Random distractors overestimate ability, staged evidence only partially improves zero-shot LLM predictions, and wrong-terminal commitment remains frequent under a locked benchmark policy. This supports more reliable diagnostic evaluation. Future medical LLM evaluation should report final-answer accuracy together with evidence-conditioned updating and terminal commitment behavior.

\begin{credits}
\subsubsection{\discintname}
The authors declare that they have no competing interests related to this work.
\end{credits}

\bibliographystyle{splncs04}
\bibliography{references}

\end{document}